\documentclass[10pt,twocolumn,letterpaper]{article}

\usepackage{btas}
\usepackage{times}
\usepackage{epsfig}
\usepackage{graphicx}
\usepackage{amsmath}
\usepackage{amssymb}
\usepackage{balance}
\usepackage{multirow}
\usepackage{soul}
\usepackage{booktabs,caption}
\usepackage[flushleft]{threeparttable}
\usepackage{placeins}
\usepackage{siunitx}
\usepackage[pagebackref=false,breaklinks=true,letterpaper=true,colorlinks,bookmarks=false]{hyperref}
\usepackage{fancyhdr}
\usepackage[skip=2pt]{caption} 
\usepackage{enumitem}
\usepackage{xcolor}
\usepackage{tensor}
\usepackage{diagbox}
\usepackage[linesnumbered,ruled,vlined]{algorithm2e}
\newcommand\Perms[2]{\tensor[^{#2}]P{_{#1}}}

\DeclareMathOperator*{\argmax}{\arg\max}

\btasfinalcopy

\begin{document}

\title{Spoofing PRNU Patterns of Iris Sensors while Preserving Iris Recognition}

\author{Sudipta Banerjee, Vahid Mirjalili, Arun Ross\\
Michigan State University\\
{\tt\small \{banerj24,mirjalil,rossarun\}@cse.msu.edu}}


\maketitle
\fancypagestyle{plain}{
	\fancyhf{} 
	\fancyhead[C]{\textcolor{red}{Published in 5th IEEE International Conference on Identity, Security and Behavior Analysis (ISBA), (Hyderabad, India), January 2019.}}
	\fancyfoot[R]{}
	\renewcommand{\headrulewidth}{0pt}
	\renewcommand{\footrulewidth}{0pt}
}

\begin{abstract}
The principle of Photo Response Non-Uniformity (PRNU) is used to link an image with its source, i.e., the sensor that produced it. In this work, we investigate if it is possible to modify an iris image acquired using one sensor in order to spoof the PRNU noise pattern of a different sensor. In this regard, we develop an image perturbation routine that iteratively modifies blocks of pixels in the original iris image such that its PRNU pattern approaches that of a target sensor. Experiments indicate the efficacy of the proposed perturbation method in spoofing PRNU patterns present in an iris image whilst still retaining its biometric content.
\end{abstract}

\section{Introduction}

The process of automatically determining the sensor that produced a given image is referred to as {\em sensor identification}. While a number of sensor identification methods have been discussed in the literature~\cite{Relwork2, Geradts_SPIE_01, Bayram_ICIP_05}, the ones based on Photo Response Non-Uniformity (PRNU)~\cite{Lukas_TIFS_06, Lukas_TIFS_08, Jess_ISP_09} have gained prominence in the recent literature. PRNU refers to the non-uniform response of individual pixels across the sensor array to the same illumination as a consequence of manufacturing defects introduced during sensor production. PRNU manifests itself as a noise pattern in the images generated by a sensor. This noise pattern is believed to be unique to every sensor~\cite{UniquePRNU}. A number of schemes have been designed to compute the PRNU noise of a sensor based on the images generated by it~\cite{SPN_Ref1_2017}. 

More recently, the principle of  PRNU has been used to perform sensor identification in the context of iris biometrics by processing the near-infrared (NIR) ocular images acquired by typical iris sensors~\cite{Uhl_2012_Baseline2,Kalka_CVPRW_15, Uhl0_IWBF_14,Uhl1_IWBF_15,Uhl3_IWBF_16,Ross_IWBF_17,IrisPRNU_18,SPN_Ref2_2017,Vatsa_18}. In this case, sensor identification (or device identification) can be used in conjunction with biometric recognition to authenticate both the identity of a device (e.g., a smartphone) as well as the individual using the device~\cite{Galdi_PRL_15}. 

Given the forensic value of PRNU in determining the origin of an image (\ie, the sensor or device that produced it), we explore if it is possible to alter an image such that its source, as assessed by a PRNU estimation scheme, is confounded. We impose two constraints:
\begin{enumerate}
\item The modified image must spoof the PRNU pattern of a pre-specified target sensor.
\item The biometric utility of the modified image must be retained, viz., the modified ocular image must match successfully with the original image.
\end{enumerate}
\begin{figure}[t]
\begin{center}
   \includegraphics[width=0.95\linewidth]{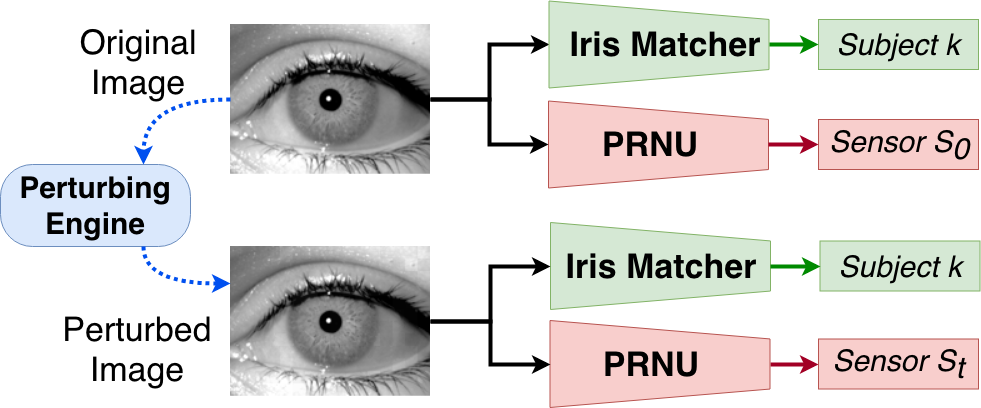}
\end{center}
   \caption{The objective of this work is to perturb an ocular (iris) image such that its PRNU pattern is modified to spoof that of another sensor, while not adversely impacting its biometric utility.}
\label{fig:paperoutline}
\end{figure}

\begin{figure*}[t]
\begin{center}
  \includegraphics[width=0.95\linewidth]{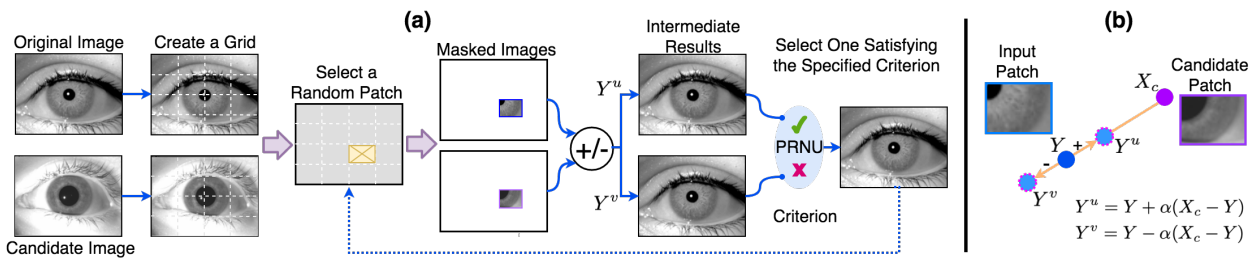}
\end{center}
   \caption{The proposed algorithm for deriving perturbations for the input image using the candidate image. (a) Steps involved in modifying the original image from the source sensor using a candidate image from the target sensor (see Algorithm~\ref{alg:find-candidate}), and (b) role of the candidate image in the perturbation engine (see Algorithm~\ref{alg:perturbations}).} 
\label{fig:algorithm-graphic}
\end{figure*}
This kind of attack can be considered as a `targeted attack', since the sensor whose PRNU pattern has to be spoofed is pre-specified. In the literature, it is also referred to as fingerprint-copy attack~\cite{Ref_Fingerprintcopy2, Uhl_2012_Baseline2}, because the objective is to copy the sensor pattern or `fingerprint' corresponding to the target sensor to an image acquired using a different source sensor. The proposed work has two distinct benefits. Firstly, it allows us to assess the feasibility of PRNU spoofing from a counter-forensic perspective. The widespread use of forensic techniques for examining the validity and origin of digital media~\cite{CounterImageForensics_Ref1, CounterImageForensics_Ref2} necessitates the study of attacks that can potentially undermine the performance of such forensic methods. For example, an adversary may maliciously attempt to link an image to a different camera in an effort to mislead law enforcement investigators~\cite{Ref_Fingerprintcopy2}. Secondly, establishing the viability of such spoof attacks would promote the development of more robust PRNU estimation schemes~\cite{SPN_Ref2_2017}. In addition, effective methods to detect such attacks can be developed if the process of spoofing is better understood. Figure~\ref{fig:paperoutline} summarizes the objective of this work.

The remainder of the paper is organized as follows. Section~\ref{PRNU} briefly reviews the PRNU based sensor identification scheme used in this work. Section~\ref{RelatedWork} presents methods that have been described in the literature for sensor anonymization and spoofing. Section~\ref{ProposedMethod} describes the proposed method for spoofing PRNU patterns. Section~\ref{Expts} provides details about the datasets used, the experimental protocols employed, and reports the results obtained using the proposed method. Section~\ref{Summary} summarizes the paper and indicates future work.

\section{Photo Response Non-Uniformity (PRNU)}
\label{PRNU}
PRNU estimation entails computing the {\em reference pattern} of a sensor based on a set of training images acquired using the sensor. This reference pattern is then used by a sensor classifier to identify the sensor that was used to acquire a given test image. This is accomplished by correlating the reference pattern of the sensor with the {\em noise residual} of the test image to compute a correlation score. The image is assigned to the sensor whose reference pattern yields the highest correlation value. Here, we used Normalized Cross-Correlation (NCC) for computing the correlation score~\cite{Uhl1_IWBF_15, Ross_IWBF_17}. PRNU estimation can be done using numerous approaches~\cite{Lukas_TIFS_06, Li_TIFS_10, Kang_TIFS_12, Li2_TIFS_16, Li3_SPL_16, SPN_Ref1_2017}. In this work, we used the Maximum Likelihood Estimation (MLE) based PRNU estimation scheme~\cite{Lukas_TIFS_08}, which has been demonstrated to suppress image artifacts not associated with the sensor-specific pattern and has resulted in very good performance~\cite{Uhl0_IWBF_14, Uhl1_IWBF_15}. 

MLE based PRNU estimation uses a weighted averaging of the noise residuals extracted from a set of training images pertaining to the sensor; each noise residual is weighted by its corresponding training image, to derive the maximum-likelihood estimate of the reference pattern. Wiener filtering and zero-mean operations are applied to the noise residuals to address interpolation artifacts arising due to the Bayer pattern. In our experiments, $L_2-$normalization of the test noise residual is performed to account for the variations in the PRNU strength of different sensors~\cite{Uhl5_IJCB_17}. The MLE of the reference pattern corresponding to a sensor is computed as, $\mathbf{\hat{K}} = \frac{\sum_{i=1}^{N} \mathbf{w}_i\mathbf{I}_i}{\sum_{i=1}^{N} \mathbf{I}_i^2}.$
Here, $\mathbf{w}_i$ is the noise residual obtained using a wavelet-based denoising filter applied to training image $\mathbf{I}_i$ and $\mathbf{w}_i = \mathbf{I}_i - F (\mathbf{I}_i)$, where, $F$ denotes the Daubechies Quadrature Mirror Filter~\cite{Lukas_TIFS_06}.

\section{Perturbing the PRNU Pattern}
\label{RelatedWork}
The counter-forensics literature describes techniques that can be used to suppress or perturb the PRNU pattern embedded in an image. This is often referred to as \textit{source anonymization}~\cite{PRNU_Perturb1_2014}, \ie, obscuring the `fingerprint' of the source sensor in an image so as to anonymize the origin of the image. Source anonymization can be used as a privacy preservation scheme, particularly relevant when the sensor-specific details can be used to associate a sensor with its owner. Assuming that each device is typically associated with a single user, device identification can be indirectly used to reveal the identity of the person possessing that specific device~\cite{AnonRef_2011}. There have been primarily two approaches to perturb the PRNU pattern for this purpose, namely, (i) compression and filtering based schemes, which typically use strong filtering schemes such as, flat-field subtraction~\cite{PRNU_attack2} or Wiener filtering~\cite{PRNU_attack4_2013} that can degrade the PRNU pattern leading to incorrect source attribution; and (ii) geometric perturbation based schemes such as `seam carving'~\cite{PRNU_attack3_2014,PRNU_attack4_2013} that distorts the alignment between the sensor reference pattern and the test noise residual, thereby impeding the process of correlating the reference pattern with the test noise residual.

In contrast to source anonymization, \textit{PRNU spoofing} not only suppresses the fingerprint of the source sensor, but it also inserts the fingerprint of the target sensor. An adversary may tamper with the digital evidence to maliciously exculpate a guilty person or worse, incriminate an innocent person. In recent literature, PRNU spoofing has been performed by two methods, namely, (i) PRNU injection and (ii) PRNU substitution. The first method adds the weighted reference pattern of a pre-selected target sensor to the input image, $\mathbf{I}$~\cite{Ref_Fingerprintcopy2}. The modified image becomes $\mathbf{I'}=[\mathbf{I}+\mathbf{I}\times\gamma \mathbf{\hat{K}_{T}}]$. Here, $\mathbf{\hat{K}_{T}}$ is the reference pattern of the target sensor $T$ and $\gamma$ is a scalar parameter. The second method subtracts the PRNU pattern of the source sensor in an image and then adds the PRNU pattern of a target sensor~\cite{PRNU_attack5_2010}. The modified image is represented as $\mathbf{I'}=\mathbf{I} - \gamma \mathbf{\hat{K}_{S}} + \beta \mathbf{\hat{K}_{T}}$. $\mathbf{I}$ belongs to the source sensor $S$, whose reference pattern is $\mathbf{\hat{K}_{S}}$. $\gamma$ and $\beta$ are scalar terms. We will use the two methods described above, \ie, PRNU injection and PRNU substitution, as baseline algorithms for comparative evaluation. The first method will be referred to as Baseline 1 and the second method will be referred to as Baseline 2. Both baseline algorithms have been shown to be successful on images acquired using commercial cameras that employ RGB sensors.

In~\cite{Uhl_2012_Baseline2}, the authors examine the viability of PRNU spoofing via injection in the context of \textit{iris sensors} operating in the NIR spectrum~\cite{PRNU_attack_NEW}. In their work, they computed the forged image as $\mathbf{I'}=[\textit{F}(\mathbf{I}) + \gamma \mathbf{\hat{K}_{T}}]$. Here, $F(\cdot)$ is the wavelet based denoising filter discussed in Section~\ref{PRNU}, and $\gamma$ is a scalar parameter. The authors further performed the triangle test to detect the spoof attack, but did not analyze the impact of the PRNU spoofing on iris recognition performance.  

In this paper, our objective is to perform PRNU spoofing in a principled manner, that works for any arbitrary pair of source and target iris sensors. In addition, we wish to retain the biometric utility of the PRNU-spoofed image. The task of spoofing can be potentially accomplished through different techniques, an example will be the use of adversarial networks that have been successfully utilized for perturbing images in the current literature~\cite{SAN,mirjalili_semi_2018}. However, a significant bottleneck of deep-learning based techniques is the need for large amount of training data for driving the perturbation process. We will demonstrate the success of the proposed PRNU spoofing scheme using small number of images ($<$1000).

\section{Proposed method}
\label{ProposedMethod}
In this section, we formally describe the objective and the method used to address this objective.

\subsection{Problem formulation}
Let $X$ denote an NIR iris image of width $w$ and height $h$, and $\mathbf{S} = \{S_1, S_2, .., S_n\}$, denote a set of $n$ sensors. Let $\phi(X,S_i)$ be the function that computes the normalized cross-correlation (NCC) between the noise residual of $X$ and the PRNU reference pattern of sensor $S_i$. Then, the sensor label for the input iris image $X$ can be determined using $\displaystyle \arg \max_{i}\{\phi(X,S_i)\}$. Furthermore, let $M$ be a biometric matcher where $M(X_1, X_2)$ determines the match score between two iris samples $X_1$ and $X_2$. Given an input iris image $X$ acquired using sensor $S_o$, a candidate image $X_c$ from the target sensor $S_t$, and an iris matcher $M$ our goal is to devise a perturbation engine $\Psi$ that can modify the input image as $Y = \Psi(X,X_c)$ such that $\phi(Y,S_o)<\phi(Y,S_t)$, and thereby predict $S_t$ as the sensor label of the perturbed image $Y\!$, while the iris matcher, $M\!$, will successfully match $Y$ with $X$. As a result, the target sensor will be spoofed, while the biometric utility of the image will be retained. This implies that the match score between a pair of perturbed images $[M(Y_1, Y_2)]$ as well as that of a perturbed sample with an original sample, $[M(X_1, Y_2)]$ and $[M(Y_1,X_2)]$, are expected to be similar to the match scores between the original samples $[M(X_1,X_2)]$. The steps used to achieve this task are described next.


\begin{table*}
\centering
\caption{Specifications of the datasets used in this work.}
\label{Tab1:Datasets}
\scalebox{0.75}{
\begin{tabular}{|llccc|}
\hline
\textbf{Dataset}    & \textbf{Sensor Name (Abbreviation) }    & \textbf{Image Size} & \begin{tabular}[c]{@{}l@{}}\textbf{Number of Images Used} \\ \textbf{(Training set+Testing set)}\end{tabular}& \textbf{Number of Subjects} \\
\hline \hline
BioCOP 2009 Set I            & Aoptix Insight (Aop)     & 640$\times$480                    & 995 (55+940)                       & 100                         \\
IITD~\cite{IITD}                           & Jiristech JPC 1000 (JPC) & 320$\times$240                   & 995 (55+940)                       & 100                         \\
CASIAv2  Device2~\cite{CASv2}         & CASIA-IrisCamV2 (IC)    & 640$\times$480                   & 995 (55+940)                       & 50                          \\
IIITD Multi-spectral Periocular (NIR subset)~\cite{IIITD}  & Cogent (Cog)            & 640$\times$480                    & 588 (55+533)                       & 62                          \\
ND CrossSensor Iris 2013 Set II~\cite{ND} & LG 4000 (LG40)        & 640$\times$480                   & 615 (55+560)                       & 99     \\ 
MMU2~\cite{MMU} & Panasonic BM-ET 100US Authenticam (Pan) & 320$\times$238 & 55 (55+0) & 6\\
ND Cosmetic Contact Lens 2013~\cite{ND} & IrisGuard IG AD100 (AD) & 640$\times$480 & 55 (55+0)&4\\
WVU Off-Axis & EverFocus Monochrome CCD (Ever)~\cite{WVU} & 640$\times$480 & 55 (55+0)&7\\
CASIAv2 Device 1~\cite{CASv2} & OKI IrisPass-h (OKI) & 640$\times$480 & 55 (55+0)&3\\
CASIAv4-Iris Thousand subset~\cite{CASv4} & IrisKing IKEMB100 (IK) & 640$\times$480 & 55 (55+0)&3 \\
ND CrossSensor Iris 2013 Set I~\cite{ND} & LG 2200 (LG22)       & 640$\times$480 & 55 (55+0)&5 \\ \hline 
\end{tabular}}
\end{table*}

\begin{algorithm}
  \small{
  \caption{\label{alg:find-candidate}Selection of the candidate image.}
  \KwIn{An image $X$ from sensor $S_o$, a gallery of images $G = \{X_1, ..., X_L\}$ from the target sensor $S_t$}
  
  \KwOut{A candidate image, $X_{c}$, selected from the gallery.}
  
  Set static parameters $K=10$ (number of random patches) and $w_p=10, h_p=10$, (patch width and height).
  
  Generate a set of $K$ random patch locations $P=\{p_1, \cdots, p_K\}$, where each patch size is $h_p \times w_p$.
  
  Compute the average pixel intensity in each patch $p_k\in P$ of the input image $X$ to obtain a vector $\mathbf{v}_X$ (of size $K$).
  
  Repeat step 3 for each of the gallery images to obtain a set of vectors $\mathbf{v}_{G_i}$, where, $i = 1,\cdots,L$. The value of $L$ (the target gallery size) depends on the number of test images indicated in the fourth column in Table~\ref{Tab1:Datasets}.
  
  Compute the correlation between $\mathbf{v}_X$ and $\mathbf{v}_{G_i}$ corresponding to each gallery image to obtain a set of $L$ correlation scores.
  
  Return candidate image $X_{c} \in G$ that has the highest correlation, \ie $X_{c} = X_{f}$ where $f = \displaystyle \argmax_{i\in [1,\cdots,L]} \{Corr(\mathbf{v}_X,\mathbf{v}_{G_i})\}$.}
     
\end{algorithm}

\subsection{Deriving perturbations and PRNU Spoofing}
\label{Derivingperturbations}
Given a single image $X$ from the source sensor $S_o$, a gallery of images $G = \{X_1, ..., X_L\}$ from the target sensor $S_t$, and a set of $K$ random patch locations $P=\{p_1, ..., p_K\}$, we first select a candidate image, $X_c$, $c\in [1,\cdots,L]$, from the gallery to perturb the input image. The candidate image is selected from the gallery such that it is maximally correlated with the input image $X$. To accomplish this goal, we select 10 patches in the input image, each of size $10\times10$ (\ie, $K=10,~h_p=10,~w_p=10$ in Algorithm~\ref{alg:find-candidate}). Now, we compute the average pixel intensity in each of these patches and create a $K$-dimensional vector $\mathbf{v}_X$. Next, for each of the $L$ gallery images, we create $\mathbf{v}_{G_i}$ where $i=[1, \cdots, L]$, by computing the average pixel intensity in the 10 patches selected previously in the input image. Finally, we compute the correlation between the vectors $\mathbf{v}_X$ and $\mathbf{v}_{G_i}$, and select the candidate image with the maximum correlation value. The steps for selecting the candidate image are described in Algorithm~\ref{alg:find-candidate}. 

After obtaining the candidate image $X_{c}$ from the gallery of the target sensor $S_t$, the perturbations for image $X$ are then derived with the help of $X_{c}$ as described in Algorithm~\ref{alg:perturbations}. The perturbation routine employs the following parameters: (i) $\alpha$ (the learning rate), (ii) $\eta$ (the termination criterion), and (iii) $m$ (the maximum number of iterations). Initially, the output perturbed image $Y^{(0)}$ is identical to the input image $X$. Next, we select a random patch location from $Y^{(0)}$, and create a mask matrix, $\mathit{Mask}$, of the same size as $Y^{(0)}$, such that the elements in $\mathit{Mask}$ are set to 1 for the row and column indices corresponding to the selected patch location. Then, the image $Y^{(0)}$ is perturbed iteratively using pixels from the same patch location in $X_c$. In each iteration, the pixels inside the selected patch are updated along two directions. The candidate image guides the direction of perturbation~\cite{vahidRef}. In the first case the perturbation is along a positive direction (implemented using line 9 in Algorithm~\ref{alg:perturbations}), which generates $Y^u$. The other direction corresponds to a negative perturbation (see line 11 in Algorithm~\ref{alg:perturbations}), which produces $Y^v$. Figure~\ref{fig:algorithm-graphic}(b) illustrates the role of the candidate image in the perturbation routine. Next, the noise residuals extracted from $(Y^u,Y^v)$ are correlated with the reference pattern of the target sensor. The perturbed image yielding the maximum correlation value is then selected as the seed image for the next iteration, $iter$. This process is repeated until the relative difference between the NCC values of perturbed image $Y^{iter}$ with respect to target sensor $S_t$ and the original sensor $S_o$ exceeds 10\%, \ie, $\eta=0.1$, or the maximum number of iterations is reached. The parameters employed in the perturbation routine are selected intuitively; for example, the learning rate is set to a small value $\alpha=0.01$ because our objective is to perturb the image while preserving its biometric utility.

\begin{algorithm}
\small{
  \caption{\label{alg:perturbations}Spoofing PRNU pattern.}
  \KwIn{An image $X_{h\times w}$ from sensor $S_o$, a candidate image $X_{c}$ from sensor $S_t$, a function $\phi(X,S_i)$ that returns the NCC value when image $X$ is correlated with the PRNU pattern of sensor $S_i$ ($i\in\{o,t\}$).}
  \KwOut{perturbed image $Y$.}
  Set static parameters $\alpha=0.01$ (learning rate), $\eta = 0.1$ (threshold), $m = 3000$ (maximum number of iterations) and $h_p=10, w_p=10$ (patch size).
  
  Initialize $iter = 0$ and $Y^{(0)} = X$.
  
 \Repeat{$\displaystyle \frac{\phi(Y^{(iter)}, S_t) - \phi(Y^{(iter)}, S_o)}{\phi(X, S_o)} > \eta$}{
   \nl -- Choose a random patch location $\left(p_x,p_y\right)$ from  $[0,\frac{h}{h_p}]$ and $[0,\frac{w}{w_p}]$ such that $0\le p_x<\frac{h}{h_p},0 \le p_y < \frac{w}{w_p}$.
   
   \nl -- Construct the mask matrix $\mathit{Mask}$ such that $\mathit{Mask}[i,j]=1$ if $\left(\lfloor \frac{i}{h_p} \rfloor , \lfloor \frac{j}{w_p} \rfloor \right) = \left(p_x,p_y\right)$, and $\mathit{Mask}[i,j]=0$ elsewhere.
   
   \nl -- Create a perturbed image in the positive direction $Y^{u} = Y^{(iter)} + \alpha \mathit{Mask}\odot \left(X_c - Y^{(iter)}\right)$. 
   
   \nl -- Create a perturbed image in the negative direction $Y^{v} = Y^{(iter)} - \alpha \mathit{Mask}\odot \left(X_c - Y^{(iter)}\right)$.
   
   \nl -- Compute the NCC values of $Y^u$ and $Y^v$ for the target sensor $S_t$, $\phi(Y^{u},S_t)$ and $\phi(Y^{v},S_t)$, respectively.
   
   \nl -- Set $Y^{(iter+1)} = Y^u$ if $\phi(Y^{u},S_t) > \phi(Y^{v},S_t)$, otherwise set $Y^{(iter+1)} = Y^v$,
   
   \nl -- $iter++$,
   
   \nl -- If $iter>m$ break the loop.
}

  Return the final perturbed image, $Y^{(iter)}.$}

\end{algorithm}

\begin{figure*}
\begin{center}
  \includegraphics[width=0.85\linewidth]{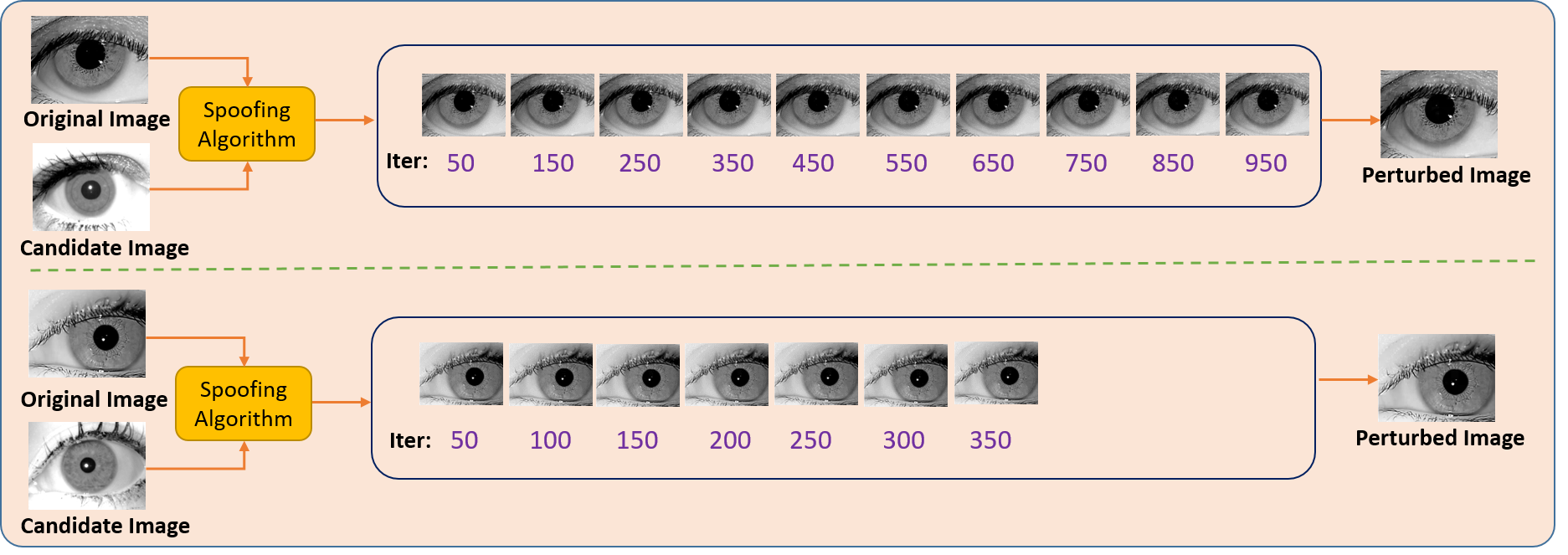}
\end{center}
  \caption{Illustration of PRNU spoofing using images belonging to the source sensor JPC and the candidate images belonging to the target sensor Aoptix.}
\label{fig:Illus_moreeg}
\end{figure*}

At the end of the routine, the perturbed image will have to be incorrectly attributed to $S_t$ by the sensor classifier. The steps of the PRNU spoofing algorithm are illustrated in Figure~\ref{fig:algorithm-graphic}(a). The sequence of modified images undergoing the perturbation routine is illustrated for two example iris images in Figure~\ref{fig:Illus_moreeg}.

\section{Experiments and Results}
\label{Expts}
In this section, we describe the datasets and sensors employed in this work, followed by the experiments conducted on the datasets. Results are reported and analyzed in the context of PRNU spoofing and iris recognition.

\subsection{Datasets}
Experiments are conducted using 11 different sensors from 11 iris datasets. The PRNU spoofing process typically involves a single source sensor and a single target sensor from the set of 11 sensors. The sensor details and image specifications of the 11 sensors are described in Table~\ref{Tab1:Datasets}. Thus, there can be a total of $\Perms{2}{11} =110$ combinations for PRNU spoofing. However, for the sake of brevity, we performed 20 different PRNU spoofing experiments involving 5 sensors: $\{Aop, JPC, IC, Cog, LG40\}$. From the set of 5 sensors listed above, each sensor serves as the source sensor while the remaining 4 sensors serve as target sensors one at a time, thus resulting in 20 different PRNU spoofing experiments.   

\begin{table*}[h]
\centering
\caption{Confusion matrix for sensor identification involving unperturbed but resized images. The test noise residuals of images from 5 sensors are compared against reference patterns from 11 sensors. The last column indicates sensor identification accuracy.}
\label{ConfMat_Ori}
\scalebox{0.85}{
\begin{tabular}{|l|lllllllllll||l|}
\hline
  \diagbox[width=5em]{\scriptsize{Actual}}{\scriptsize{Predicted}}  & Aop          & JPC          & IC           & Cog          & \small{LG 40} & Pan&AD&Ever&OKI&IK&   \small{LG 22}       & \begin{tabular}[c]{@{}l@{}}Accuracy \\(\%)\end{tabular} \\  \hline
Aop & \textbf{900} & 1            & 2            & 1            & 9 &4&3&3&9&7&1          & 95.74                               \\
JPC  & 2            & \textbf{919} & 4            & 2            &5 &0&0&4&1&2&1            & 97.77                               \\
IC & 0            & 0            & \textbf{940} & 0            & 0 &0&0&0&0&0&0           & 100                                 \\
Cog & 2            & 1            & 2            & \textbf{546} & 2 &0&2&0&0&5&0           & 97.51                               \\
LG40  & 0            & 0            & 0            & 0            & \textbf{529} & 0&0&3&1&0&0 & 99.25  \\ \hline
\end{tabular}}
\end{table*}


\subsection{Sensor identification before PRNU spoofing}
Due to variations in image size of the source and target sensors, all images were resized to a fixed spatial resolution of 160 $\times$ 120 to facilitate PRNU spoofing. We then evaluated the sensor identification accuracy based on these resized images prior to PRNU spoofing. This is to determine if resizing impacts sensor identification accuracy. The sensor identification involves deriving sensor reference patterns using 55 training images, as used in~\cite{Ross_IWBF_17} from each of the 11 sensors, followed by extraction of test noise residuals from images belonging to the 5 sensors, and finally correlating them. The subjects in the training set and the test set are disjoint. The sensor identification accuracy and the corresponding confusion matrix is presented in Table~\ref{ConfMat_Ori}. The results indicate a very high sensor identification accuracy using the MLE PRNU scheme on the resized images. So we use the resized images in the experiments below.

\subsection{Sensor identification after PRNU spoofing}
\label{B1B2_Reference}
The PRNU spoofing process involves perturbing the original image from a source sensor using a candidate image belonging to the target sensor, whose PRNU needs to be spoofed. The impact of the perturbations on spoofing the PRNU pattern has been reported in terms of \textit{Spoof Success Rate} (SSR), which computes the proportion of test images from the source sensor classified as belonging to the target sensor after perturbing using Algorithm~\ref{alg:perturbations}. The results of spoofing are presented in Table~\ref{Tab:PerturbAop}.

\begin{figure}[h]
\centering
\includegraphics[width=0.7\linewidth]{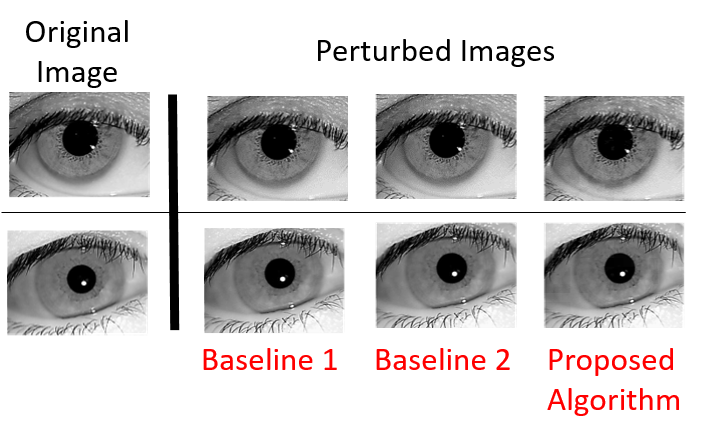}
\caption{\label{fig:demoperturbed}Example of PRNU spoofed images originating from the JPC 1000 sensor (first column) is illustrated for Baseline 1 (second column), Baseline 2 (third column) and the proposed method (last column). Here, the target sensor is Aoptix.} 
\end{figure}

We implemented Baseline 1 and Baseline 2 algorithm described in Section~\ref{RelatedWork}. Baseline 2 is implemented following normalization of the source and target reference patterns with respect to the maximum intensity of the PRNU present in the two reference patterns. The normalization is required to account for the variation in the PRNU strength associated with different sensors. Ideally, the scalar terms $\gamma$ and $\beta$, which serve as parameters in the baseline algorithm, need to be optimized through grid-search for a specific pair of source ($S_o$) and target ($S_t)$ sensors. However, we set the scalars to a static value of 1 for two reasons: (i) for ease of computation and (ii) to provide fair comparison with the proposed algorithm which also uses fixed values of parameters for all pairs of sensors. The baseline algorithms are state-of-the art to the best of our knowledge and are, therefore, used for comparative evaluation. Examples of perturbed outputs of images spoofed using Baseline 1, Baseline 2, and the proposed algorithm are presented in Figure~\ref{fig:demoperturbed}. 


\begin{table*}[t]
\centering
\caption{Results of PRNU spoofing where the target sensors (along the second column) are spoofed by perturbing the images from 5 source sensors, namely, Aop, JPC, IC, Cog and LG40 (along the first column). The test noise residual after the perturbation process is compared against the reference patterns of 11 sensors (see Table~\ref{Tab1:Datasets}). The last 3 columns indicate the proportion of the perturbed images successfully classified as belonging to the target sensor and is denoted as the Spoof Success Rate (SSR). The highest values of the SSR are bolded.}
\label{Tab:PerturbAop}
\scalebox{0.8}{
\begin{tabular}{|l|l|lllllllllll|c|c|c|}
\hline
\begin{tabular}[c]{@{}l@{}}Original\\ Sensor\end{tabular} & \begin{tabular}[c]{@{}l@{}}Target \\ Sensor\end{tabular} & \multicolumn{11}{c|}{Sensor classes compared against perturbed PRNU} & \begin{tabular}[c|]{@{}l@{}}SSR (\%) for \\proposed method \end{tabular} & \begin{tabular}[c|]{@{}l@{}}SSR (\%) for \\Baseline 1\end{tabular} & \begin{tabular}[c|]{@{}l@{}}SSR (\%) for \\Baseline 2\end{tabular}\\ \hline
\multirow{5}{*}{Aop}                                      &                                                          & Aop  & JPC  & IC   & Cog & LG40 & Pan & AD & Ever & OKI & IK & LG22 &  & &                                                                \\ \cline{3-13}
                                                           & JPC                                                      & 4    & \textbf{894}  & 3    & 3   & 8    & 2   & 2  & 2    & 9   & 12 & 1    &\textbf{ 95.11} & 92.55   &67.98                                                \\
                                                          & IC                                                       & 21   & 0    & \textbf{891}  & 0   & 6    & 2   & 1  & 5    & 6   & 5  & 3  & \textbf{94.79}  & 92.77 &13.51                                                             \\
                                                         
                                                          & Cog                                                      & 7    & 2    & 3    & \textbf{890} & 7    & 5   & 2  & 2    & 13  & 5  & 4  &   \textbf{94.68}   & 79.89   &0.21                                                        \\
                                                          & LG40                                                     & 66   & 4    & 4    & 4   & \textbf{836}  & 3   & 0  & 4    & 7   & 8  & 4  & \textbf{88.94}  & 79.15  &10.00  \\ \hline \hline
                                                           \multirow{4}{*}{JPC}                                      
                                                          & Aop                                                       & \textbf{905}   & 18    & 3  & 3   & 4   & 0   & 0  & 2    & 2   & 2  & 1  &   \textbf{96.28}   & 49.15  &1.91                                                        \\
                                                          & IC                                                      & 2    & 209  & \textbf{712}    & 2   & 4    & 2   & 1  & 3    & 2   & 2 & 1  & 75.74   & 99.79   &\textbf{100}                                                         \\
                                                          & Cog                                                      & 3    & 94    & 4    & \textbf{817} & 5    & 5   & 0  & 5    & 2  & 1  & 4  &  \textbf{86.91  }   & 35.53 &0.21                                                        \\
                                                          & LG40                                                     & 1   & 61    & 3    & 1   & \textbf{861}  & 5   & 0  & 3    & 1   & 2  & 2  &  \textbf{91.60}  & 8.09  &9.26  \\ \hline \hline           
                                                          \multirow{4}{*}{IC}                                      
                                                          & Aop                                                       & \textbf{910}   & 0    & 30  & 0   & 0   & 0   & 0  & 0    & 0   & 0  & 0 &  \textbf{96.81 }     & 48.72  &0                                                       \\
                                                          & JPC                                                      & 0    & \textbf{797}  & 143    & 0   & 0    & 0   & 0  & 0    & 0   & 0 & 0  & 84.79   & \textbf{100}&53.09                                                            \\
                                                          & Cog                                                      & 0    & 0    & 243    & \textbf{697} & 0    & 0   & 0  & 0    & 0  & 0  & 0  & \textbf{74.15}   & 46.70&0                                                           \\
                                                          & LG40                                                     & 0   & 0    & 46    & 0   & \textbf{894}  & 0   & 0  & 0    & 0   & 0  & 0   &\textbf{95.11 }  & 1.91 &0.11    \\ \hline \hline

                                                           \multirow{4}{*}{Cog}                                      
                                                          & Aop                                                       & \textbf{552}   & 0    & 0  & 0   & 2   & 0   & 2  & 0    & 0   & 4  & 0  & 98.57    & \textbf{100}&38.57                                                           \\
                                                          & JPC                                                     & 1    & \textbf{546}  & 0   & 0   & 1    & 0   & 2  & 2    & 0   & 8 & 0   & 97.50  & \textbf{100}&\textbf{100}                                                            \\
                                                          & IC                                                      & 2    & 0    & \textbf{545}    & 2 & 2    & 0   & 2  & 0    & 1  & 5  & 1  & 97.32   & \textbf{100}&\textbf{100}                                                            \\
                                                          & LG40                                                     & 1   & 0    & 0    & 0   & \textbf{550}  & 0   & 2  & 1    & 0   & 6  & 0  & \textbf{98.21}  & 82.32 &35.00    \\ \hline \hline   
                                                           \multirow{4}{*}{LG40}                                      
                                                          & Aop                                                       & \textbf{330}   & 0    & 3  & 0   & 198   & 0   & 0  & 0    & 1   & 1  & 0  & \textbf{61.91}  & 9.94&1.31                                                             \\
                                                          & JPC                                                      & 0    & \textbf{491}  & 0    & 0   & 38    & 0   & 0  & 2    & 1   & 1 & 0  &\textbf{92.12}  & 9.38 &24.20                                                            \\
                                                          & IC                                                      & 0    & 0    & \textbf{393}    & 0 & 136    & 0   & 0  & 3    & 1  & 0  & 0   & 73.73  & 11.44&\textbf{99.44}                                                             \\
                                                          & Cog                                                     & 0   & 0    & 0    & \textbf{479}   & 50  & 0   & 0  & 2    & 1   & 1  & 0  &\textbf{89.87}   & 4.69 &0.19       \\ \hline \hline
                                                          \multicolumn{13}{|l|}{\textbf{Average SSR (\%)}}& \textbf{89.21}& 57.60&32.75 \\ \hline
\end{tabular}}
\end{table*}


\textbf{Results in Table~\ref{Tab:PerturbAop} indicate that 15 out of 20 times the proposed algorithm outperforms Baseline 1 technique, and performs considerably better than Baseline 2 method 16 out of 20 times. The average SSR of the proposed algorithm outperforms the baseline algorithms by a significant margin.} We believe that the parameters $\gamma$ and $\beta$ need to be tuned accurately for each pair of source and target sensors to ensure the success of the baseline algorithms. On the other hand, the proposed algorithm is successful for static parameter values: the size of patches ($h_p\times w_p$), the threshold $\eta$, the learning rate $\alpha$, and the number of patches ($K$) (see Section~\ref{Derivingperturbations}). The PRNU is successfully spoofed by the proposed method in most of the cases barring the case where the target sensor is Aoptix and the source sensor is LG 4000 ($\approx$62\% SSR). Inspection of the images acquired using LG 4000 sensor reveals the presence of image padding, which may negatively impact the PRNU spoofing process.

\begin{figure}[h]
\centering
\includegraphics[width=0.98\linewidth]{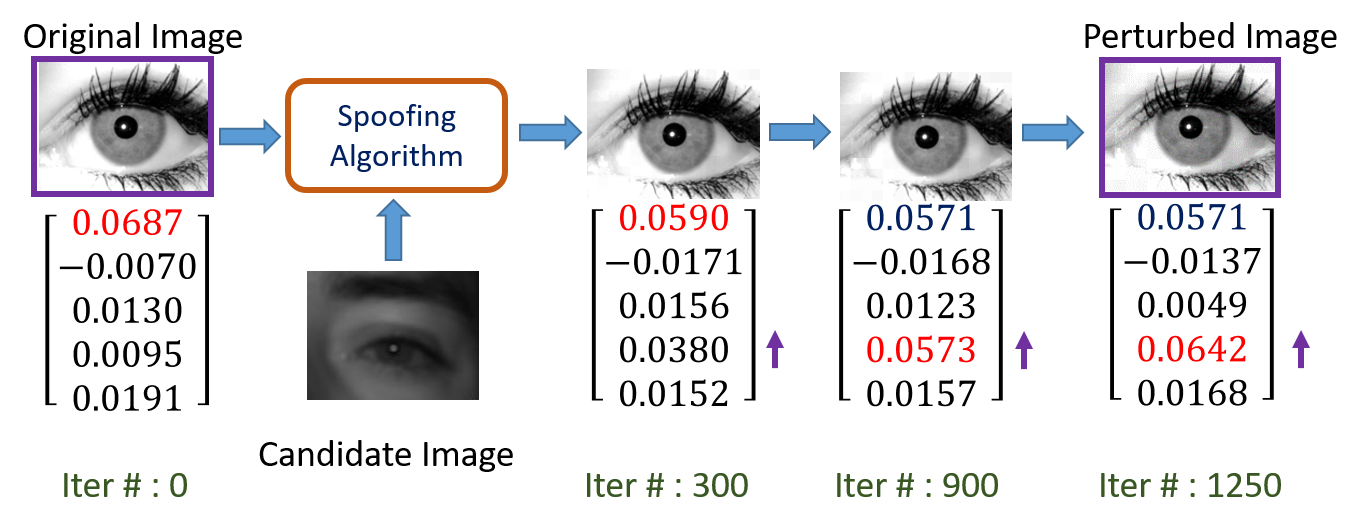}
\caption{\label{Fig:Results_Perturb}Intermediate images generated when an image from the Aoptix ($S_o$) sensor is perturbed using a candidate image from Cogent ($S_t$). For the sake of brevity, NCC values corresponding to the reference patterns of the first 5 sensors in Table~\ref{Tab1:Datasets} are mentioned in the figure. The arrows indicate the increase in the NCC values corresponding to the target sensor.} 
\end{figure}

Figure~\ref{Fig:Results_Perturb} shows an input image undergoing iterative perturbations. The original (unperturbed) image belongs to the Aoptix sensor and is perturbed using a candidate image from the target sensor, Cogent. The subsequent shift of the NCC values from being the highest for the source sensor (Aoptix) to being the highest for the target sensor (Cogent), indicates the success of the proposed method.

The average number of iterations required for successful PRNU spoofing varied between 200 to 2200. Another experiment is conducted to study the impact of increasing the number of iterations on the proposed PRNU spoofing process. This experiment is conducted for the specific case where the source sensor is LG 4000 and the target to be spoofed is the Aoptix sensor. The reason for selecting this pair is due to the poor SSR reported for this specific set of sensors (see the fifth block in Table~\ref{Tab:PerturbAop}). We speculate that with an increase in the number of iterations, the PRNU spoofing process will succeed and improve the SSR as a result. In this regard, in the new experimental set-up, the maximum number of iterations was set to 6000 (twice the earlier terminating criterion). As a result, the SSR increased considerably from 61.91\% to 79.73\%, \ie, a $\approx$ 18\% increase was observed. 425 out of 533 test images belonging to the LG 4000 sensor were successfully classified as originating from the Aoptix sensor when the number of iterations was increased. 

\subsection{Retaining biometric matching utility}
\label{IrisMatching}
The impact of the perturbations on iris recognition performance is evaluated next using the VeriEye iris matcher~\cite{VeriEye}. We designed three experiments for analyzing biometric matching performance. First, the match scores between all pairs of iris samples before perturbation were computed. In the second experiment, we computed the match scores between all pairs of perturbed samples. In the third experiment, we computed match scores between all iris samples before perturbation and all samples after perturbation. This is referred to as the cross-matching scenario. In the third set of experiments, the genuine scores are computed by employing 2 sample images (from the same subject): one sample belonging to the set of unperturbed images and the other sample from the set of perturbed images. The impostor scores are generated by pairing samples belonging to different subjects: one image is taken from the set of unperturbed images, while the other is taken from the set of perturbed images.
\begin{figure*}
\centering
\includegraphics[width=0.98\linewidth]{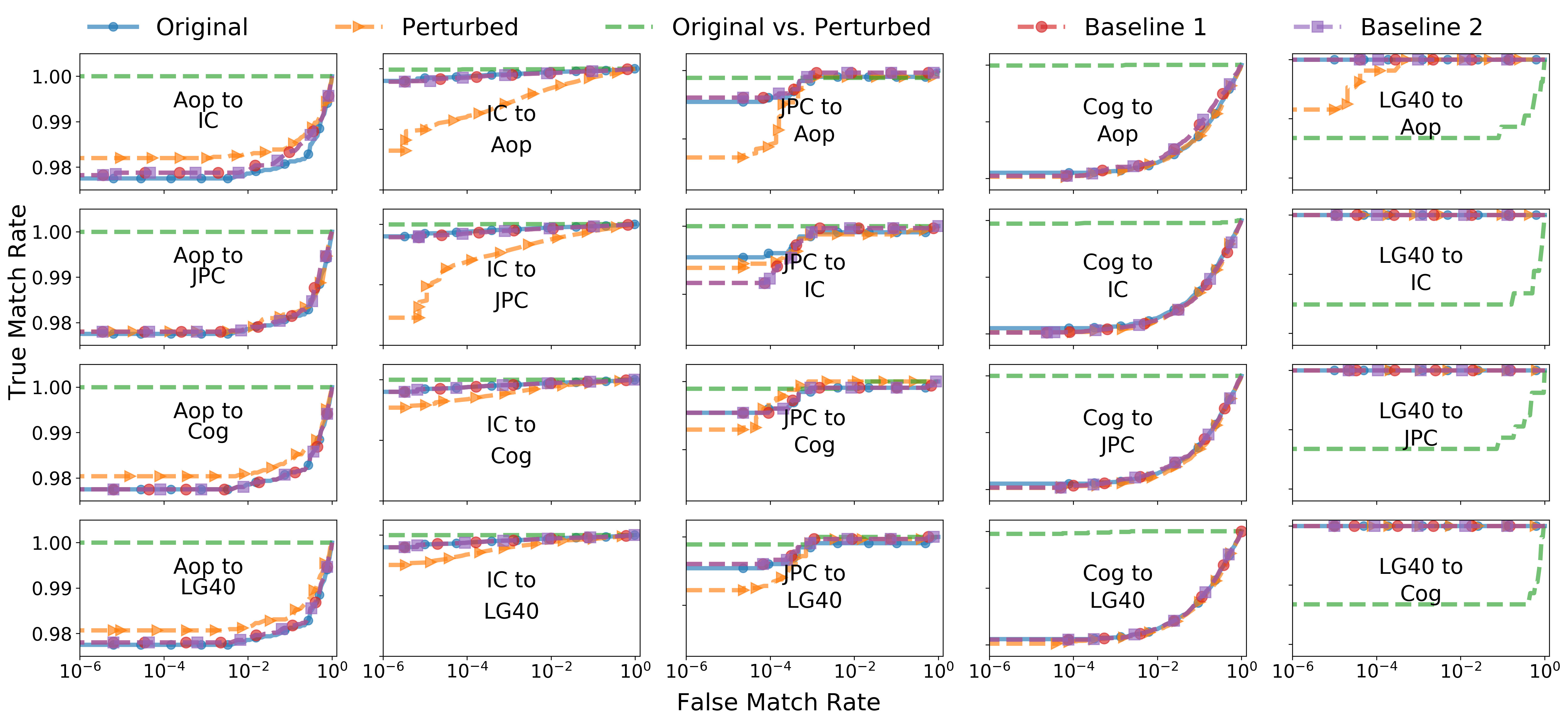}
\caption{\label{fig:roc-matching}ROC curves of matching performance obtained using the VeriEye iris matcher software. The terms `Original', `Perturbed' and `Original vs. Perturbed' indicate the three different matching scenarios (see Section~\ref{IrisMatching}). `Original' indicates matching only unperturbed images; `Perturbed' indicates matching only perturbed images; `Original vs. Perturbed' indicates the cross-matching case where unperturbed images are matched against perturbed images. Note that the curves obtained from perturbed images match very closely with the curves corresponding to the unperturbed images illustrating preservation of iris recognition for each sensor depicted in each column. The results are compared with Baseline 1 and 2 algorithms discussed in Section~\ref{B1B2_Reference}.} 
\end{figure*} 

Figure~\ref{fig:roc-matching} shows the ROC curves obtained from these three experiments. The ROC curves confirm that the perturbed images do not negatively impact the matching utility. In the case of all the sensors, the ROC curves of the perturbed images are within a $1\%$ deviation from the ROC curve of the original samples before perturbation, except for the IrisCam (IC) sensor. Further, we note that the matching performance of original samples from the Cogent (Cog) sensor is degraded to begin with. We believe the reason for this degraded performance is due to the low quality of the original images. Yet, perturbations have not further deteriorated the matching performance, as evidenced by the before- and after-perturbation ROC curves that are very similar to each other.

\begin{figure}
\centering
\includegraphics[width=0.7\linewidth]{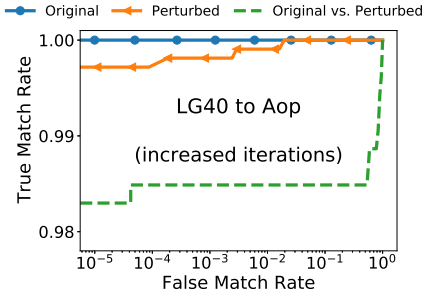} 
\caption{\label{fig:roc-matching-und-more} Impact of increase in the number of iterations on iris recognition performance for the pair of LG 4000 (source) and Aoptix (target) sensors. } 
\end{figure}

In addition, the iris recognition performance after PRNU spoofing using the baseline algorithms is analyzed. The results indicate that the proposed method is comparable to the baseline algorithms in terms of iris recognition performance. Furthermore, we conducted a fourth experiment, where we analyzed the matching performance of those LG4000 iris images that were perturbed to spoof the Aoptix sensor after increasing the number of iterations. The result confirms that increasing the number of iterations to improve the SSR does not degrade matching performance, as is evident in Figure~\ref{fig:roc-matching-und-more}.

\paragraph{In summary, the following salient observations in the context of both PRNU spoofing and iris recognition preservation can be made.}

\begin{itemize}
 \item The PRNU pattern of a sensor can be successfully spoofed by \textit{directly} modifying an input image, without invoking the sensor reference pattern of the target sensor. Experiments are conducted using 11 iris sensors, and the PRNU spoofing process is demonstrated using 5 sensors and compared with existing approaches. Results show that the proposed spoofing method outperforms Baseline 1 by 31.6\% and Baseline 2 by 56.4\% in terms of average spoof success rate. 
 \item The proposed spoofing algorithm uses identical parameters, such as the size of patches and learning rate for all pairs of source and target sensors. This obviates the need to fine tune the method for different pairs of sensors.
 \item The iris recognition performance of the images perturbed using the proposed algorithm is retained within 1\% of the original. This suggests the success of the proposed spoofing method in retaining the biometric utility of the modified images.
 \end{itemize}

\section{Summary and Future Work}
\label{Summary}
In this work, we design a method for PRNU spoofing that preserves biometric recognition in the context of NIR iris images. In the proposed strategy, a test image belonging to a particular sensor is modified iteratively using patches from a candidate image belonging to a target sensor, whose PRNU is to be spoofed. We examine the impact of these perturbations on PRNU spoofing as well as iris recognition performance. Experiments are conducted in this regard using 11 sensors and compared with two existing PRNU spoofing algorithms. Results show that the proposed method can successfully spoof the PRNU pattern of a target sensor and does not significantly impact the iris recognition performance in a majority of the cases.
\par
Future work will involve testing the proposed PRNU spoofing process on a larger set of sensors and analyzing the impact of the number of candidate images on the spoof success rate. The iterative spoofing routine can be expedited by perturbing multiple image patches simultaneously (instead of one patch at a a time). Finally, we will look into developing new sensor identification schemes that are resilient to spoof attacks as well as methods to detect such attacks.   

\section*{Acknowledgement} 
We would like to thank Denton Bobeldyk and Steven Hoffman from the iPRoBe Lab for sharing their iris recognition code. This material is based upon work supported by the National Science Foundation under Grant Number $1618518$.

{\small
\bibliographystyle{ieee}
\balance
\bibliography{egbib}
}

\end{document}